\documentclass[10pt,twocolumn,letterpaper]{article}

\usepackage{iccv}              

\definecolor{iccvblue}{rgb}{0.21,0.49,0.74}
\usepackage[pagebackref,breaklinks,colorlinks,allcolors=iccvblue]{hyperref}
\usepackage{booktabs}

\def\paperID{002} 
\def\confName{ICCV}
\def\confYear{2025}

\title{ArtFace: Towards Historical Portrait Face Identification via Model Adaptation}

\author{
Francois Poh, Anjith George and Sébastien Marcel\\
Idiap Research Institute\\
Rue Marconi 19, CH - 1920, Martigny, Switzerland\\
{\tt\small francois.poh22@imperial.ac.uk, \{anjith.george, marcel\}@idiap.ch}
}

\begin{document}
\maketitle
\section{Introduction}
\label{sec:intro}

Identifying sitters in historical paintings is a key task for art historians, offering insight into their lives and how they chose to be seen~\cite{salavessa2024proto}. However, the process is often subjective and limited by the lack of data and stylistic variations. Automated facial recognition is capable of handling challenging conditions and can assist, but while traditional facial recognition models perform well on photographs~\cite{huang2008labeled}, they struggle with paintings due to domain shift and high intra-class variation~\cite{gupta2018deep}. Artistic factors such as style, skill, intent~\cite{huber2022verification, salavessa2024proto}, and influence from other works~\cite{artuk2024copying} further complicate recognition.

The early work of Srinivasan et al.~\cite{srinivasan2015face} approached the problem using local and anthropometric facial features to statistically assess similarity between portrait pairs. More recent approaches rely heavily on deep learning, particularly convolutional neural networks (CNNs). Gupta et al.~\cite{gupta2018deep} used Siamese CNNs to compare controversial historical portraits, leveraging style transfer techniques to augment training data while preserving essential facial content. Building on this, Huber et al. ~\cite{huber2022verification} fine-tuned a ResNet-100 model on an augmented dataset and introduced a confidence measure by estimating uncertainty in face comparison scores. Salavessa et al. ~\cite{salavessa2024proto} combined deep learning with handcrafted facial feature ratios, using a VGG-based approach.

In this work, we investigate the potential of foundation models to improve facial recognition in artworks. By fine-tuning foundation models and integrating their embeddings with those from conventional facial recognition networks, we demonstrate notable improvements over current state-of-the-art methods. Our results show that foundation models can bridge the gap where traditional methods are ineffective. Paper page at \footnote{\url{https://www.idiap.ch/paper/artface/}}.
\paragraph{Contributions}
We address the task of sitter identification through the following contributions:
\textbf{(i)} Fine‑tuning the vision–language foundation model CLIP~\cite{radford2021learning} with LoRA~\cite{hu2022lora} on portrait face images;
\textbf{(ii)} Complementing it with embeddings from an adapted face recognition network (AntelopeV2/IResNet100~\cite{deng2019arcface}); and
\textbf{(iii)} Fusing the embeddings via a simple yet effective normalised concatenation strategy.
Comprehensive experiments on the public \emph{Historical Faces} dataset~\cite{huber2022verification} show that our method reduces baseline error rates and outperforms prior approaches.

\begin{figure*}[!htb]
    \centering
    \includegraphics[width=0.75\linewidth]{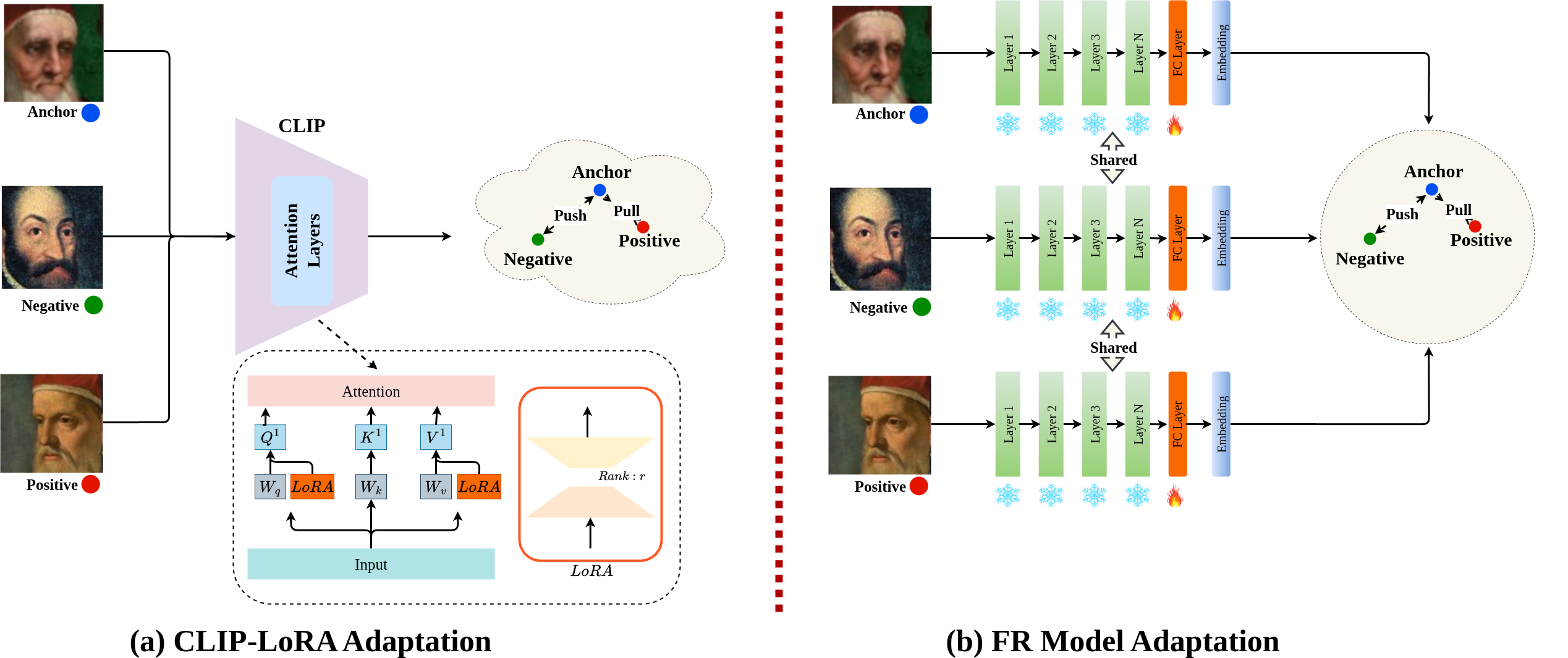}
    \caption{Overview of the proposed method: (a) LoRA-based adaptation of the CLIP model, and (b) head adaptation using triplet loss.}
    \label{fig:framework}
\end{figure*}
\section{Proposed Method}
\label{sec:proposed}

\textbf{CLIP Foundation Model}: CLIP is a vision–language foundation model trained on 400 million image–text pairs collected from the internet~\cite{radford2021learning}. It learns to align images and text in a shared embedding space, enabling zero-shot transfer to a wide range of tasks such as image classification and visual understanding. Since foundation models are trained on diverse data, they capture broad concepts, detect stylistic differences, and can use contextual information relevant to paintings that traditional facial recognition overlooks. Prior work has shown that combining CLIP embeddings with those from facial recognition models yields modest improvements on photographic data~\cite{sony2025foundation}, and this benefit can extend to painted portraits as well. For our experiments, we use the CLIP ViT-B/16 model from OpenAI, selected for its balance between tunability and computational efficiency.

\textbf{Face Recognition Model}: AntelopeV2 is a state-of-the-art face recognition (FR) model built on the IResNet100 architecture~\cite{deng2019arcface} and trained on the large-scale Glint360k dataset. It achieves high accuracy and robustness across challenging variations in pose, age, and lighting, making it a strong and widely adopted baseline in face verification benchmarks. In addition to these models, we also use a Commercial off-the-shelf face recognition model as another baseline (COTS).

\textbf{Model Adaptation}: Both the CLIP and face recognition (FR) models are adapted using portrait images to better align with the domain of historical paintings, as shown in Fig. \ref{fig:framework}. For CLIP, we apply Low-Rank Adaptation (LoRA)~\cite{hu2022lora}, inserting LoRA layers into the query (Q) and value (V) matrices of the attention mechanism. This approach introduces low-rank updates, allowing efficient fine-tuning with significantly fewer trainable parameters and reduced memory usage, while maintaining performance. For the IResNet100 model, adaptation is minimal; only the final linear layer is fine-tuned to adjust the embeddings for the new domain.

\textbf{Fusion}: Embeddings from the tuned and untuned IResNet100 models, along with the tuned CLIP model, are first individually normalised, then concatenated and re-normalised. This process effectively performs feature fusion, combining complementary information from each model.

\section{Experiments}
\label{sec:experiments}

\textbf{Dataset}: The Historical Faces dataset~\cite{huber2022verification} consists of 766 paintings of 210 different sitters. The paintings are preprocessed by detecting facial key points and aligning and cropping to 112x112 for both IResNet100 and CLIP. This was split for training and evaluation by identity in a 60:20:20 ratio for the train, validation, and test set, respectively.

\textbf{Training Details}: The CLIP-LoRA model is trained using triplet loss~\cite{schroff2015facenet} with cosine distance and a margin of 0.5, with a batch size of 48. Training employs early stopping with a patience of 10 epochs to prevent overfitting. Hard negative mining is performed at each epoch: 30\% of the negative samples are selected from the top 50 hardest negatives per anchor, while the remaining 70\% are randomly drawn from the next 450 hardest examples. The model is optimised using Adam with a learning rate of 1e-5. LoRA layers, with rank 16, are applied to the query (Q) and value (V) matrices in all transformer layers, implemented using the PEFT library~\cite{peft}. IResNet100 uses the same method and parameters for training, but no LoRA layers are applied; instead, only tuning the final linear layer. The final output score is the cosine similarity between the concatenated embeddings.

\subsection{Experimental Results}

\begin{figure*}[!htb]
    \centering
    \includegraphics[width=0.75\linewidth]{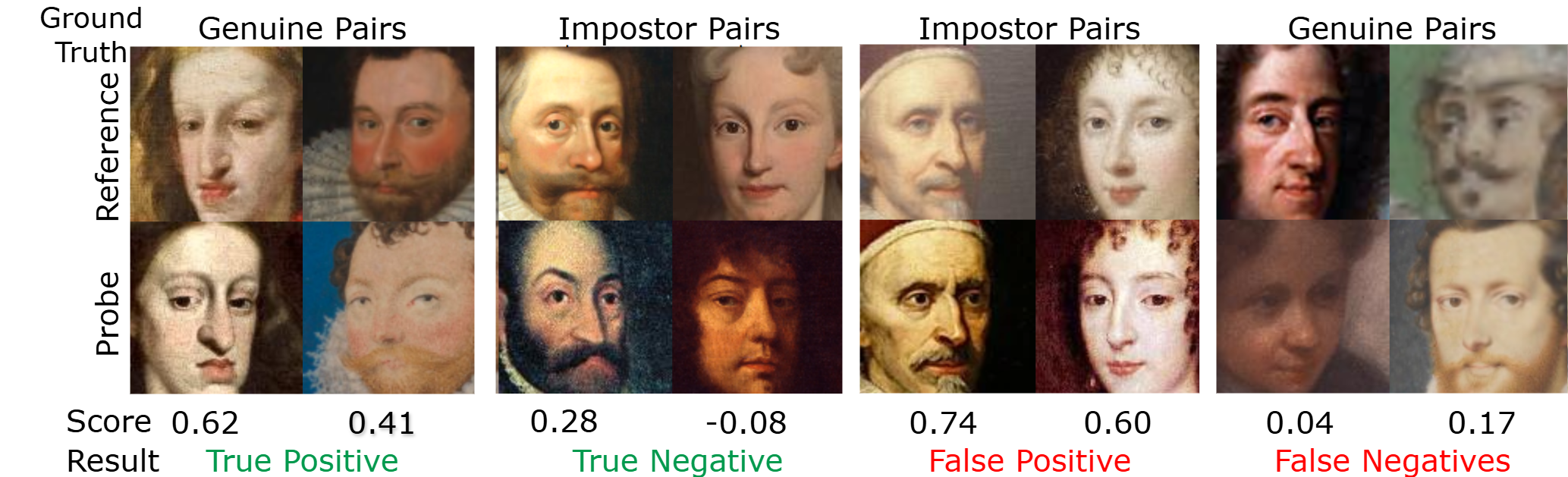}
    \caption{Examples of successful and failed comparisons. Each pair shows the reference and probe images, with the associated cosine similarity score. High similarity scores indicate successful matches, while low scores represent mismatches. }
    \label{fig:success_failure}
\end{figure*}

\textbf{Results with IResNet100 Tuning}: Table \ref{tab:iresnettuning} demonstrates that tuning IResNet100 primarily enhances TAR\footnote{True Accept Rate (TAR) = 1 - False Non-Match Rate (FNMR)} at low FAR\footnote{False Accept Rate (FAR) = False Match Rate (FMR)} values ($<$1\%), with only modest improvements in EER\footnote{Equal Error Rate (EER) = Error Rate when FMR=FNMR}. While fusion performs slightly worse than the tuned model in low-FAR TAR, it consistently outperforms both the base and tuned models overall.

\begin{table}[htb]
    \centering
    \caption{Performance of IResNet100 variants with and without tuning and fusion.}
    \label{tab:iresnettuning}
    \resizebox{0.49\textwidth}{!}{
    \begin{tabular}{lccc}
    \toprule
        \textbf{Model} & \textbf{EER$\downarrow$} & \textbf{TAR@0.1\%FAR$\uparrow$} & \textbf{TAR@1\%FAR$\uparrow$} \\
        \midrule
        IResNet100-Base                & 14.0\% & 29.9\%          & 55.1\% \\
        IResNet100-Tuned               & 13.5\% & \textbf{36.5\%} & 53.7\% \\
        IResNet100-Fusion (Base+Tuned) & 13.5\% & 33.6\%          & \textbf{58.4\%} \\
    \bottomrule
    \end{tabular}
    }
\end{table}

\textbf{Results with CLIP-LoRA Tuning}: In this section, we describe the fine-tuning of CLIP-LoRA using triplet loss, both with and without hard negative mining. All other relevant parameters remain unchanged from previous experiments. Although the hard negative mining variant does not achieve higher TAR at all FAR levels, it shows improved performance above approximately 1\% FAR. As shown in Table \ref{tab:cliptune}, this approach also leads to better EER.

\begin{table}[]
    \centering
    \caption{Performance of loss functions for tuning CLIP. Models with Hard Negative Mining (HN) perform better overall.}
    \label{tab:cliptune}
    \resizebox{0.49\textwidth}{!}{
    \begin{tabular}{lccc}
    \toprule
        \textbf{Model} & \textbf{EER $\downarrow$ } & \textbf{TAR@0.1\%FAR$\uparrow$} & \textbf{TAR@1\%FAR$\uparrow$} \\
        \midrule
        CLIP-Base                    & 17.9\% & 8.4\%  & 33.2\% \\
        CLIP-LoRA(Triplet, HN)      & 13.1\% & 17.8\% & 43.5\% \\
        CLIP-LoRA(Triplet, w/o HN)  & 13.9\% & 16.8\% & 43.9\% \\
    \bottomrule
    \end{tabular}
    }
\end{table}

\begin{figure}
    \centering
    \includegraphics[width=0.8\linewidth]{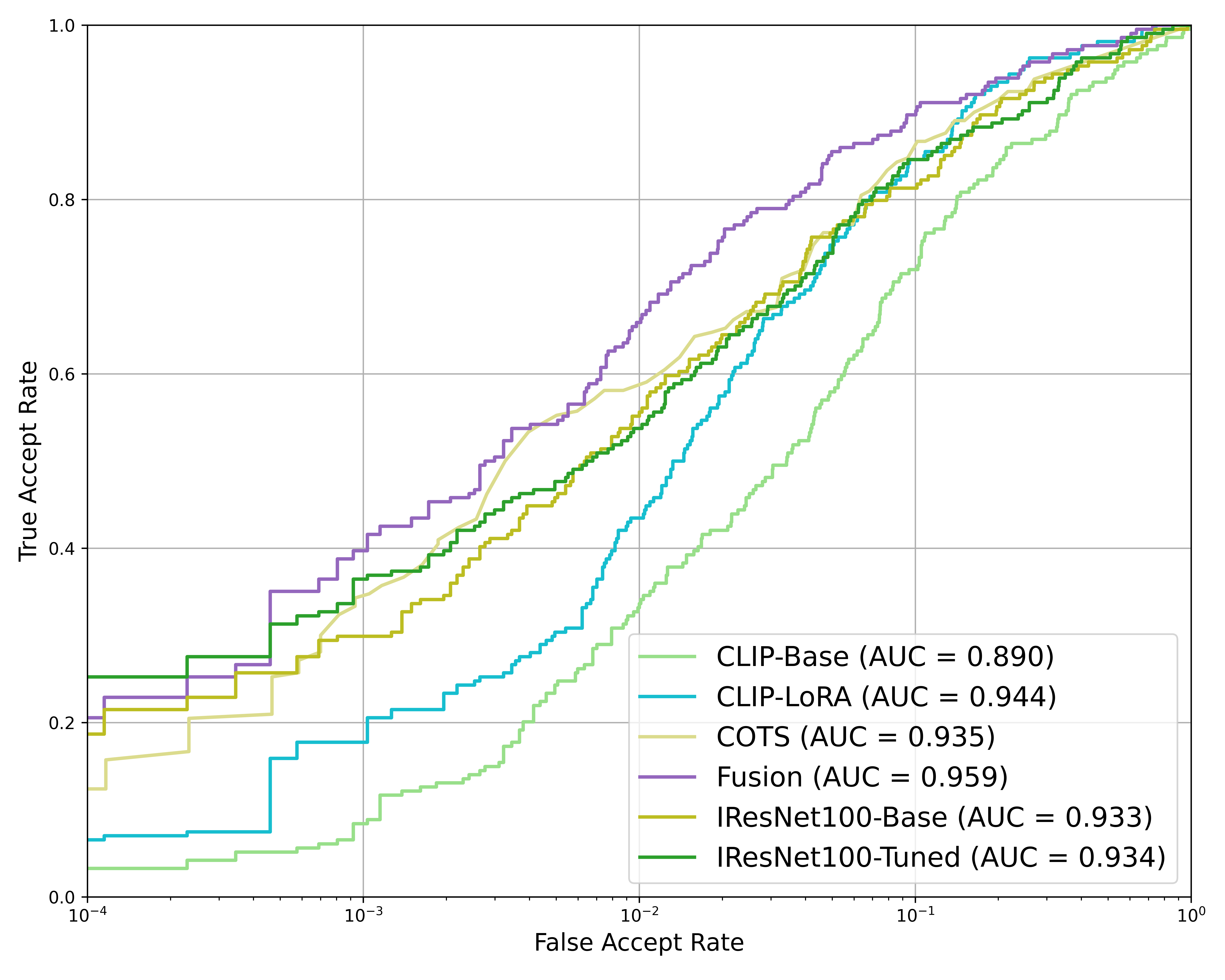}
    \caption{ROC curves of tuned and base CLIP, IResNet100, COTS and proposed fusion method. Fusion provides consistent improvements even at low FAR.}
    \label{fig:roc}
\end{figure}

\textbf{Results with Fusion}: Table \ref{tab:all} compares the performance of fusion methods with individual IResNet100 and CLIP models on the Historical Faces test split. Although the base CLIP model underperforms compared to IResNet100, fusion leads to improvements in EER. Fine-tuning CLIP-LoRA further enhances performance across all four metrics, with additional gains observed when the tuned IResNet100 is included in the fusion. The corresponding ROC curves are presented in Fig. \ref{fig:roc}.

\begin{table}[]
    \centering
    \caption{Performance Comparison of Base, Tuned models, Fusion, and COTS FR Systems. Fusion enhances overall accuracy.}
    \label{tab:all}
    \resizebox{0.49\textwidth}{!}{
    \begin{tabular}{lccc}
    \toprule
        \textbf{Model} & \textbf{EER$\downarrow$} & \textbf{TAR@0.1\%FAR$\uparrow$} & \textbf{TAR@1\%FAR$\uparrow$} \\
        \midrule
        COTS FR system               & 12.6\% & 34.3\% & 58.1\% \\
        \hline
        CLIP-Base                    & 17.9\% & 8.4\%  & 33.2\% \\
        IResNet100-Base              & 14.0\% & 29.9\% & 55.1\% \\
        \hline
        CLIP-Base + IResNet100-Base  & 13.1\% & 29.0\% & 54.7\% \\
        CLIP-Base + IResNet100-Tuned & 12.6\% & 35.1\% & 57.9\% \\
        CLIP-LoRA + IResNet100-Base  & 11.1\% & 34.6\% & 62.6\% \\
        CLIP-LoRA + IResNet100-Tuned & 10.7\% & 39.7\% & 62.15\% \\
        \midrule
        \textbf{CLIP-LoRA + IResNet100-Base + IResNet100-Tuned} & \textbf{9.9\%} & \textbf{39.7\%} & \textbf{65.9\%} \\
    \bottomrule
    \end{tabular}
    }
\end{table}

\textbf{Discussions}: Tuning IResNet100 on the Historical Faces dataset yields only modest gains, but fusing tuned and untuned versions outperforms either alone, suggesting combining representations can help, even within a single architecture.

CLIP-LoRA tuning shows triplet loss with hard negative mining performs best, significantly improving over base CLIP. Other loss functions underperform, likely due to limited data, especially in terms of images per identity.

Fusion experiments show that combining foundation models with conventional face recognition networks improves performance across all evaluation metrics, especially in low false acceptance rate (FAR) scenarios. This indicates that foundation models capture valuable information from portraits that conventional networks may overlook.

\textbf{Visualisations}: To better understand both successful and failed matches, we show sample images along with their similarity scores in Fig.~\ref{fig:success_failure}. Genuine comparisons can fail in challenging cases, particularly when there are significant differences in artistic style or medium, sitter age, and often compounded when multiple factors overlap. Conversely, some impostor comparisons yield surprisingly high similarity scores. This typically occurs when artists base their work on earlier portraits or when sitters resemble each other and are depicted in a similar style or by the same artist.

\section{Conclusions}
\label{sec:conclusions}

In this work, we show that lightweight tuning of vision–language foundation models~\cite{hu2022lora}, combined with domain-adapted face recognition networks, can effectively bridge the domain gap between photographs and paintings. Our fusion approach achieves state-of-the-art accuracy in sitter identification. Face recognition on artworks remains a particularly difficult task compared to traditional FR due to the scarcity of labelled data, stylistic variation, and the interpretive nature of portraiture. However, the results show that adapting modern architectures to this setting is feasible and promising. This opens up new research avenues, including synthetic data generation~\cite{george2025digi2real} to augment the limited training set and heterogeneous domain adaptation techniques~\cite{george2024modality} to improve generalisation across visual domains.

\section{Acknowledgements}
This research was supported by the InterArt project, funded by Loterie Romande.

\clearpage
{
    \small
    \bibliographystyle{ieeenat_fullname}
    \bibliography{main}
}

\end{document}